\documentclass[sigconf]{acmart}
\AtBeginDocument{%
  }

\setcopyright{acmlicensed}
\copyrightyear{2018}
\acmYear{2018}
\acmDOI{XXXXXXX.XXXXXXX}
\acmConference[Conference acronym 'XX]{Make sure to enter the correct
  conference title from your rights confirmation email}{June 03--05,
  2018}{Woodstock, NY}
\acmISBN{978-1-4503-XXXX-X/2018/06}

\usepackage{xcolor}
\usepackage{enumitem}
\usepackage{amsfonts}
\usepackage{subcaption}
\usepackage{booktabs}
\usepackage{amsmath}
\usepackage{amsthm}
\usepackage{eucal}
\usepackage{algorithm}
\usepackage{algorithmic}
\usepackage{float}
\usepackage{threeparttable}

\begin{document}

\title{Advancing Multi-Organ Disease Care: A Hierarchical Multi-Agent Reinforcement Learning Framework}

\author{Daniel Jason Tan}
\email{djtan@u.nus.edu}
\affiliation{%
  \institution{National University of Singapore}
  \city{Singapore}
  \country{Singapore}
}

\author{Qianyi Xu}
\email{
e0673214@u.nus.edu}
\affiliation{%
  \institution{National University of Singapore}
  \city{Singapore}
  \country{Singapore}
}

\author{Dilruk Perera}
\email{
dilruk@nus.edu.sg}
\affiliation{%
  \institution{National University of Singapore}
  \city{Singapore}
  \country{Singapore}
}

\author{Kay Choong See}
\email{
kaychoongsee@nus.edu.sg}
\affiliation{%
  \institution{National University of Singapore}
  \city{Singapore}
  \country{Singapore}
}

\author{Mengling Feng}
\email{
ephfm@nus.edu.sg}
\affiliation{%
  \institution{National University of Singapore}
  \city{Singapore}
  \country{Singapore}
}

\renewcommand{\shortauthors}{Trovato et al.}

\begin{abstract}
% Multi-organ system diseases present significant challenges due to their simultaneous impact on multiple organ systems, necessitating complex and adaptive treatment strategies. Despite recent advancements in AI-powered clinical decision support systems, existing solutions are limited to individual organ systems. They often ignore the intricate dependencies among organ systems and are thereby limited in their ability to provide holistic treatment recommendations that are useful in practice. We propose a novel hierarchical multi-agent reinforcement learning (HMARL) framework to address these challenges. This framework uses dedicated agents for each organ system, and model dynamics through explicit inter-agent communication channels, enabling coordinated treatment strategies across organs systems Furthermore, we introduce a dual-layer state representation technique to contextualize patient conditions at various hierarchical levels, enhancing the treatment accuracy and relevance. Through extensive qualitative and quantitative evaluations in managing sepsis—a complex multi-organ disease—our approach demonstrates its ability to learn effective treatment policies that significantly improve patient survival rates. This framework marks a substantial advancement in clinical decision support systems, pioneering a comprehensive approach for multi-organ treatment recommendations.

In healthcare, multi-organ system diseases pose unique and significant challenges as they impact multiple physiological systems concurrently, demanding complex and coordinated treatment strategies. Despite recent advancements in the AI based clinical decision support systems, these solutions only focus on individual organ systems, failing to account for complex interdependencies between them. This narrow focus greatly hinders their effectiveness in recommending holistic and clinically actionable treatments in the real world setting. To address this critical gap, we propose a novel Hierarchical Multi-Agent Reinforcement Learning (HMARL) framework. Our architecture deploys specialized and dedicated agents for each organ system and facilitates inter-agent communication to enable synergistic decision-making across organ systems. Furthermore, we introduce a dual-layer state representation technique that contextualizes patient conditions at both global and organ-specific levels, improving the accuracy and relevance of treatment decisions. We evaluate our HMARL solution on the task of sepsis management, a common and critical multi-organ disease, using both qualitative and quantitative metrics. Our method learns effective, clinically aligned treatment policies that considerably improve patient survival. We believe this framework represents a significant advancement in clinical decision support systems, introducing the first RL solution explicitly designed for multi-organ treatment recommendations. Our solution moves beyond prevailing simplified, single-organ models that fall short in addressing the complexity of multi-organ diseases.
\end{abstract}




\received{20 February 2007}
\received[revised]{12 March 2009}
\received[accepted]{5 June 2009}

\maketitle

\section{Introduction}

Multi-organ diseases are characterized by the sequential or simultaneous impairment of multiple organs \cite{asim2020multiple}. They present significant challenges in clinical management due to the difficulty in balancing therapeutic trade-offs, and their potential for life-threatening outcomes. Treating these diseases requires a holistic approach that accounts for interdependencies between different organ systems \cite{tian2023heterogeneous}. Existing guideline-based approaches treat organs in isolation and rely on one-size-fits-all recommendations \cite{whelehan2020medicine}. A recent example is COVID-19, which primarily affects the respiratory system, but could also lead to dysfunction in the immune, nervous, and gastrointestinal systems \cite{thakur2021multi,bhadoria2021multi}. Another example is sepsis, a serious condition resulting from the body’s dysregulated response to infection. Sepsis can lead to widespread inflammation, coagulation abnormalities, and metabolic disruptions, cascading into multi-organ dysfunction \cite{greco2017platelets}.

Recent advances in reinforcement learning (RL), have shown promise in optimizing clinical decision-making for complex diseases. Its capacity to learn adaptive policies from high-dimensional, complex data makes it a powerful tool for such tasks. Notably, deep RL has been applied to sepsis treatment, to learn dynamic treatment regimes (DTRs) from electronic health records (EHRs) of intensive care unit (ICU) patients \cite{komorowski2018artificial}. Subsequent work has explored model-free approaches, such as Dueling Double Deep Q-Networks (D3QN) \cite{raghu2017deep,wu2023value} and model-based approaches \cite{raghu2018model}. RL has also been applied to chronic diabetes management
\cite{wang2023optimized,zheng2021personalized,liu2020deep}. Despite the multi-organ nature of many major diseases, existing RL approaches have predominantly focused on recommending treatments targeting a single organ system at a time. For example, current solutions for sepsis primarily address the cardiovascular system via fluids and vasopressor dosing optimization \cite{liu2020reinforcement, komorowski2018artificial, raghu2017deep}. This is a significant limitation, as treatments for one organ can significantly influence the efficacy or safety of treatments for another. For example, recommending vasopressors (VAs) to stabilize blood pressure can increase renal impairment for a patient with concurrent renal dysfunction \cite{yagi2021treatment}. While solutions with explicit consideration of multiple organ systems exist, their primary focus has been diagnosis, instead of complex treatment recommendation \cite{khan2023improving,kaur2024systematic}.

Multi-organ disease treatment recommendation introduces complexities beyond what traditional RL and non-RL-based recommendation solutions can effectively manage. For example, as the number of organ systems or treatments increase, the number of combinations of actions increases exponentially, making action spaces untenable for standard single-agent RL algorithms to navigate. Additionally, patients' measured physiological variables will relate to  each organ system's unique physiology in different ways, adding an additional layer of complexity that a multi-organ solution must account for. Consequently, there is a clear need to develop a robust and holistic treatment recommendation solution tailored for multi-organ system disease management.

We propose a hierarchical multi-agent RL (HMARL) solution, which divides the complex task of multi-organ treatment recommendation among a hierarchy of specialized sub-agents. Each agent operates within its own localized state and action spaces, simplifying decision-making and allowing agents to focus exclusively on relevant subspaces. Carefully designed inter- and intra- agent communication mechanisms enable collaboration when necessary, alleviating the burden on individual agents and leading to more efficient training and faster convergence. Moreover, understanding patient states and their dynamics within localized contexts is essential for accurate treatment recommendations. For example, when treating the cardiovascular system, more focus should be on factors such as ejection fraction and cardiac enzyme levels, whereas renal treatments require attention to factors such as glomerular filtration rate and electrolyte balances \cite{deferrari2021renal}. To achieve this, we propose a multi-layer hierarchical representation technique that first captures broad health indicators at the root level and then refines them into organ-specific representations, which are utilized by the corresponding agents at these levels. Collectively, this HMARL system, combined with the multi-layer hierarchical representation technique, effectively manages the complexities of multi-organ treatment recommendation.

We summarize our main contributions as follows:
\begin{itemize}
    \item To the best of our knowledge, we propose the first-of-its-kind multi-organ treatment recommendation solution.
    \item We introduce a compact HMARL framework that decomposes the complex task of multi-organ disease management into manageable subtasks, handled by specialized sub-agents operating within localized state and action spaces, independently and collaboratively.
    \item We develop a multi-layer hierarchical representation technique that learns broad and specific patient representations, tailored to treatment context, aiding accurate decision-making at multiple levels.
    \item We demonstrate the effectiveness of our approach through extensive experiments, showing its superiority over traditional RL models in handling multi-organ interdependencies and improving treatment outcomes.
\end{itemize}
Our offline RL-based solution is trained and tested on retrospective public data as is standard in healthcare RL to ensure safety and stability. 
% through curated data and well-defined clinical patterns. 
The solution serves purely as a clinical decision support system 
% and is not intended to take on a decision-making role, but rather 
that offers data-driven treatment recommendations as an expert-in-the-loop solution. There are no ethical violations in its development or use.
% the development or use of this solution. 

\section{Methodology}
\subsection{Hierarchical Decomposition}
\label{sec:hd}
% \begin{figure}[h!]
%     \centering
%     \includegraphics[width=0.4\textwidth]{figures/hmarl_archi_nos3.pdf}
%     \caption{HMARL solution architecture for training and coordination of treatment agents. The root agent $M_{Rt}$ selects actions from the root action space $A_{Rt}^0$, intra-organ action spaces $A_{Rt}^{Neu}, A_{Rt}^{Car}, A_{Rt}^{Ren}$, or the inter-organ mixed action space $A_{Rt}^{OMix}$. Within each organ system $o \in \{\text{Neu}, \text{Car}, \text{Ren}\}$, a master agent $M_o$ coordinates treatment selection. The neuro-only master agent $M_{Neu}$ directs sub-agents $M_{Neu}^{S1}, M_{Neu}^{S2}, M_{Neu}^{Mix}$ for specific treatments or mixed-treatment combinations. Similarly, the cardio master agent $M_{Car}$ and renal master agent $M_{Ren}$ manage their respective sub-agents, such as $M_{Car}^{IV}, M_{Car}^{VA},$ and $M_{Car}^{Mix}$. Mixed-organ coordination is achieved through the QMIX-based agent $M_{OMix}$, which integrates outputs from organ-specific agents ($M_{OMix}^{Neu}, M_{OMix}^{Car}, M_{OMix}^{Ren}$) to enable inter-organ communication. Dashed lines represent intra-organ communication, while solid lines illustrate inter-organ communication pathways.}
%     \label{fig:model_structure}
% \end{figure}

\begin{figure*}[t]
  \centering
  \begin{subfigure}[t]{0.48\textwidth}
    \centering
    \includegraphics[width=\textwidth]{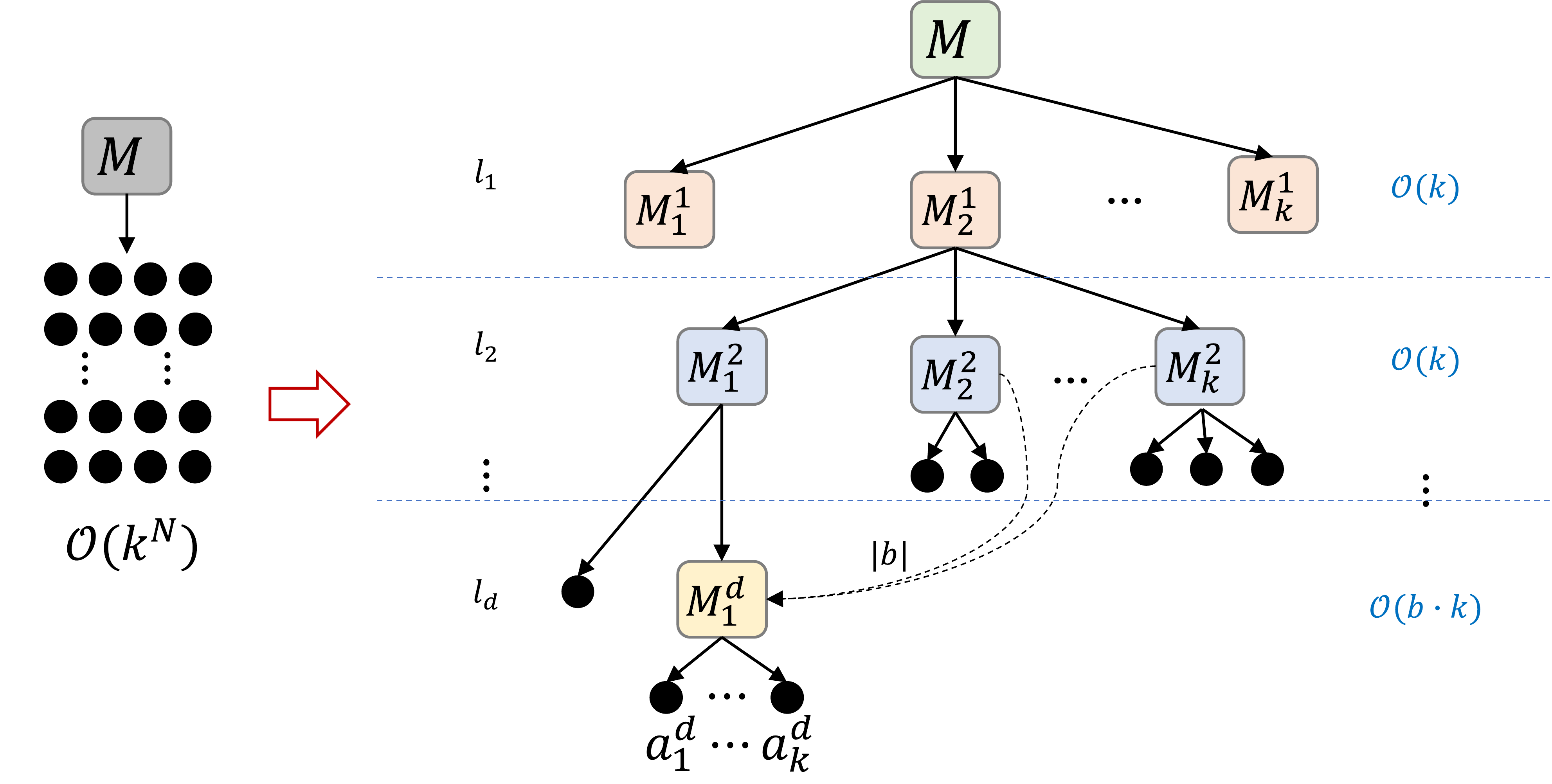}
    \caption{Hierarchical decomposition and complexity reduction. The full decision space of size $\mathcal{O}(k^N)$ is decomposed into $d$ layers, each selecting from at most $k$ options, with $b$ parallel agents per layer. Resulting complexity is reduced to $\mathcal{O}(d \cdot b \cdot k)$.}
    \label{fig:hierarchy_complexity}
  \end{subfigure}
  \hfill
  \begin{subfigure}[t]{0.48\textwidth}
    \centering
    \includegraphics[width=\textwidth]{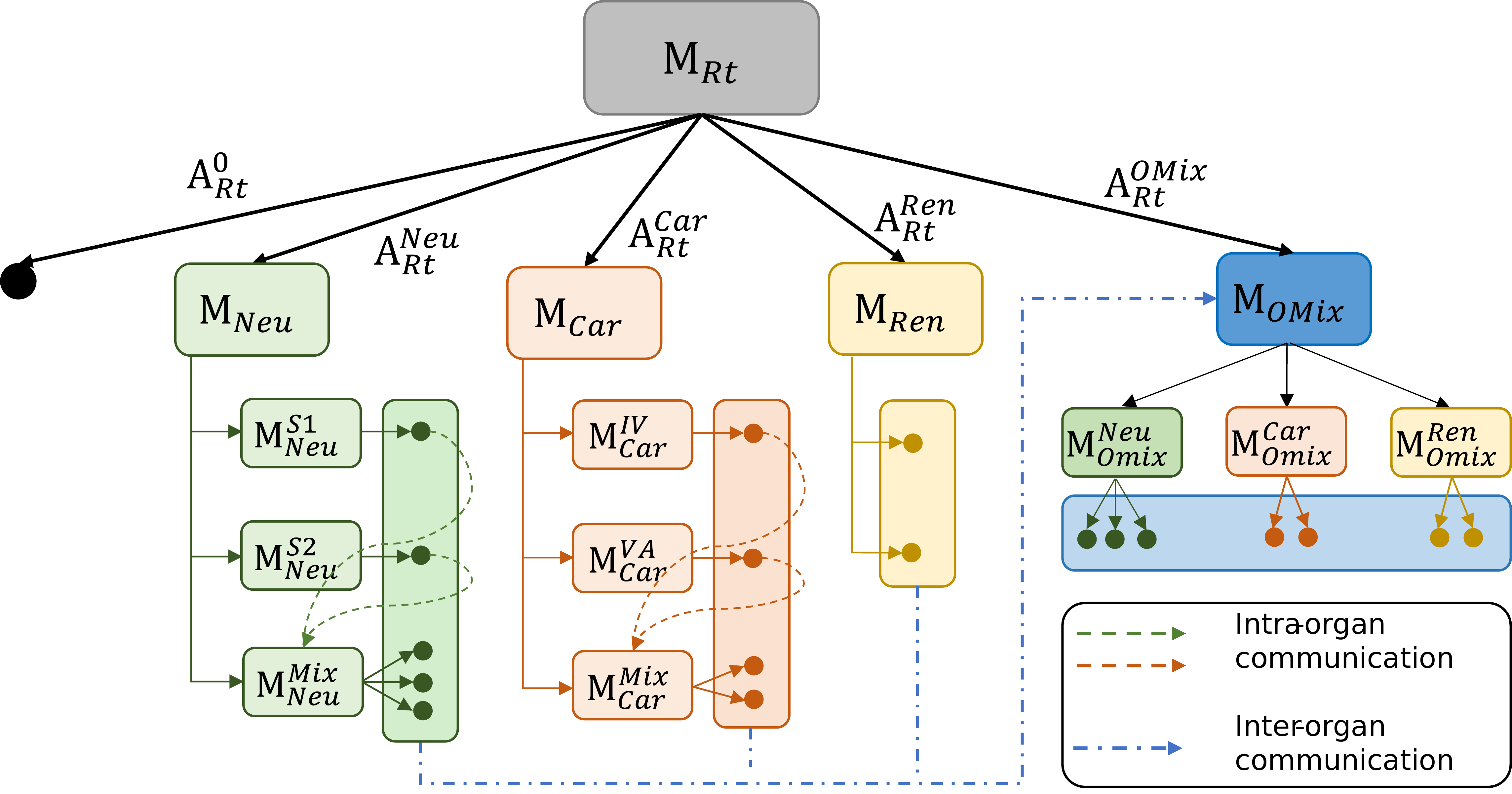}
    \caption{Instantiation for multi-organ treatment recommendation. Root agent $M_{Rt}$ delegates to organ-specific agents or a cross-organ mixture agent $M_{OMix}$. Treatment-level decisions are made by specialized sub-agents, with intra- and inter-organ communication.}
    \label{fig:hmarl_instantiation}
  \end{subfigure}
  \caption{Hierarchical multi-agent framework and its medical instantiation. (a) A generic HMARL framework achieves scalable decision-making via depth-wise decomposition. (b) Specific application to multi-organ treatment recommendation with coordinated agent behavior.}
  \label{fig:model_structure}
\end{figure*}

Complex decision-making tasks such as treatment recommendation or distributed control involves complex interaction among multiple subsystems or decision domains. They often contain large combinatorial action spaces, contextual dependencies and the need for local specialization with global coordination. To address these challenges, we propose a general hierarchical MARL framework that decomposes the overall problem into multiple payers of specialized decision makers (agents). Each layer addresses a different abstraction enabling a scalable, interpretable and modular policy learning.

\subsubsection{HMARL Framework Overview:}
Proposed HMARL framework is organized as follows. At the highest level, a root agent selects broad high-level options or task branches. These options invoke intermediate coordinator agents that manage related subtasks within specialized domains or scopes. Leaf level sub agents are responsible for fine-grained decisions such as selecting specific actions to perform within their specialized action subspaces. Importantly, we include both intra-branch (local) and inter-branch (global) communications modules to share key information to coordinate decisions across boundaries, where necessary. This hierarchical decomposition help reduce the burden on individual agents through smaller manageable scopes and supports structured collaboration through communication pathways and shared latent representations. The framework is flexible, supporting both independent and cooperative training regimes, and scalable to complex and heterogeneous action spaces (i.e., discrete, continuous and hybrid).  

\subsubsection{Scalability and Complexity:}
Consider a decision problem with \(N\) treatment dimensions (e.g., organ or treatment modality), each with a local action space \(\mathcal{A}_i\) of size \(|\mathcal{A}_i| = k\). A standard flat RL agent operates on the full joint space \(\mathcal{A} = \prod_{i=1}^{N} \mathcal{A}_i\), resulting in an exponential decision complexity of ${\mathcal{O}(k^N)}$. In contrast, proposed hierarchical MARL framework decomposes this problem across a hierarchy of depth \(d\), where decisions are made sequentially by specialized agents. At each layer, an agent selects from at most \(k\) options. Some agents (e.g., mixture agents) may invoke up to \(b\) sub-agents in parallel to gather auxiliary recommendations. Under this setting, the per-episode decision complexity is reduced to $\mathcal{O}(d \cdot b \cdot k)$ (see Figure \ref{fig:hierarchy_complexity}). This formulation assumes conditional branching, where only relevant sub-agents are activated. If each agent (or layer) addresses on average \(\overline{m}\) decision dimensions, the minimum hierarchy depth satisfies $d \geq \left\lceil \frac{N}{\overline{m}} \right\rceil, \quad \text{with} \quad b \leq \min(d, N)$. 

Theoretically, \(d\) could linearly scale with \(N\) yet it is substantially smaller in practice. For example, in our instantiation below with \(N=6\) treatment dimensions, a depth of \(d=3\) is sufficient, as the decision space is structured around natural clinical groupings. Even in the worst-case scenario (\(d = N\)), complexity is \(\mathcal{O}(N \cdot b \cdot k)\), which is polynomial and far more tractable than the exponential complexity. Moreover, each agent operates within a localized subspace \(\mathcal{A}_i \subset \mathcal{A}\) and exclusively trained on relevant samples. This reduces gradient interference from unrelated subtasks, improves sample efficiency, and mirrors real-world collaborative clinical practice. The architecture supports plug-and-play extensibility. New treatment modalities or decision dimensions can be integrated by adding new agents without retraining the full hierarchy, allowing the overall system to scale gracefully with added complexity.

\subsubsection{Clinical Use Case:} 
Proposed HMARL framework is domain-agnostic and can generate multiple decompositions for a given task, with varied effectivenesses based on the selected hierarchy and communication pathways. Guided by our clinical collaborators, we instantiate the framework analogous to a coordinated team of medical specialists and supervisors at practice. The model considers seven treatments across three organs (see Figure \ref{fig:hmarl_instantiation}). This decomposition restricts each subtask to a defined subspace, significantly reducing the decision making complexity while enabling selective coordination when necessary.

At the top level, the root agent \(M_{Rt}\) selects one of five high-level options: No-treatment (\(\mathcal{A}_{Rt}^{0}\)), Neuro-only (\(\mathcal{A}_{Rt}^{Neu}\)), Cardio-only (\(\mathcal{A}_{Rt}^{Car}\)), Renal-only (\(\mathcal{A}_{Rt}^{Ren}\)), or Organ mixture (\(\mathcal{A}_{Rt}^{OMix}\)). Using the commonly selected \textit{no-treatment} option at this level reduces the burden on downstream agents by pruning unnecessary branches early and reducing overall action space. When an organ-specific action is selected, the corresponding domain-level agent (i.e., \(M_{Neu}\), \(M_{Car}\), or \(M_{Ren}\)) is invoked. Neuro and Cardio agents operate similarly where each chooses between individual treatments (i.e., S1 or S2 for Neuro; IV or Vaso for Cardio) or a mixture, invoking subsequent sub-agents to recommend treatment dosages. Intra-organ communication is used for treatment mixtures (\(\mathcal{A}_{Neu}^{Mix}\), \(\mathcal{A}_{Car}^{Mix}\)) to effectively aggregate recommendations from respective single-treatment sub-agents, enhancing local coordination. In contrast, the Renal agent \(M_{Ren}\) directly selects among four diuretic levels or dialysis, as treatment mixing is not applicable. Multi-organ decisions (\(\mathcal{A}_{Rt}^{OMix}\)) are taken by the coordinator agent \(M_{OMix}\) using the inter-organ communication to obtain their treatment recommendations, reducing the need to learn these from scratch. Accordingly, it invokes a combination of sub-agents (\(M_{OMix}^{Neu}\), \(M_{OMix}^{Car}\), \(M_{OMix}^{Ren}\)), each of which incorporates relevant signals from the other domains. For instance, \(M_{OMix}^{Neu}\) considers recommendations from Cardio and Renal sub-agents. This cross-communication allows agents to adjust outputs to account for interactions, streamlining coherent multi-organ plans. Agents are trained only on samples drawn from their respective action subspaces to improve efficiency and minimize interference. For example, \(M_{Car}^{IV}\) learns from IV-only cases, while \(M_{Car}^{Mix}\) is trained on samples where both IV and Vaso were used.

This instantiation reflects a real-world clinical setting where a lead physician delegates to specialists. The resulting hierarchy decomposes decision-making across dedicated and mixture agents, enabling scalable inter- and intra- organ coordination. This task-specific decomposition and structured communication protocols supports modular policy learning making it highly applicable to broader multi-organ treatment scenarios.

\subsection{RL Components}
\subsubsection{States:}
% We used the patient state representation method from \cite{perera2023demystifying}. This method combines current and contextual patient state embeddings by using a two-layer approach. The first layer encodes raw physiological features into structured latent representations, and the second layer models higher-order interactions between embeddings. It not only reduces state space cardinality and complexity and allows for end-to-end learning of effective embeddings.
% \begin{figure}[H]
%     \centering
%     \includegraphics[width=0.5\textwidth]{figures/state_representation.png}
%     \caption{State Representation}
%     \label{fig:state_representation}
% \end{figure}

Accurate patient understanding is essential for effective multi-organ treatment recommendations, as it requires integrating complex interdependencies of physiological features and organ functions. Therefore, we propose a hierarchical patient representation approach that acknowledges the varying significance of raw features at different analytical levels.

At the root level, \textit{Unified State Representations} are learned to extract broad health indicators and their dynamics, providing a foundational understanding of patients. This representation is used for broader decision-making at the root level, and sets the stage for subsequent, more granular recommendations. At the organ levels, they are refined to learn \textit{Targeted State Representations}, tailored to unique physiological requirements and interrelationships of specific organs. For instance, in cardiac treatments, the embeddings prioritize ejection fraction and cardiac enzyme levels, capturing cardiac health indicators. In contrast, renal treatments focus on glomerular filtration rate and electrolyte balances, assessing renal function. This approach ensures that organ-specific recommendations are precise and relevant.

This dual-layer hierarchical representation strategy balances broad applicability with detailed specificity, enhancing decision-making considering each organ's condition 
% within the overall health context. 
It e comprises of the following levels:
 
\noindent\textbf{\textit{Unified State Representations:}} Inspired by the representation learning proposed in \cite{perera2023demystifying} we learn the unified representations as follows. Each patient at time $t$ is represented by their raw $d$-dimensional feature, $x_t = \{x_{t,1}, x_{t,2}, \ldots, x_{t,d}\} \in \mathbb{R}$ (for details on $x_t$, see Appendix Section \textit{Feature Processing} ). At the root agent level, these features are transformed using dense latent embeddings $E^{Rt} = \{e_{1}^{Rt}, e_{2}^{Rt}, \cdots, e_{d}^{Rt}\}\in \mathbb{R}^{d\times k}$. Each $k$-dimensional latent embedding vector $e_{i}^{Rt}\in \mathbb{R}^k$ transforms its corresponding raw feature into a more informative dense latent representation. The resultant patient-specific embeddings are represented as $F^{Rt}_t = \{f_{t,1}^{Rt}, f_{t,2}^{Rt}, \cdots, f_{t,d}^{Rt}\}\in \mathbb{R}^{d\times k}$, where $f_{t,i}^{Rt} = (x_{t,i}\cdot e_i^{Rt})$. These latent embeddings $e_i^{Rt}$ are generic at the root agent level, providing a holistic understanding of the patient by transforming each feature into a homogeneous latent space. The vector $f_{t,i}^{Rt}$ denotes the transformed representation specific to the patient at time $t$.

To capture the complex interdependencies among these features, a higher-order interaction layer is introduced. This layer computes the element-wise product between all pairs of embeddings in  $F_t^{Rt}$ resulting an interaction matrix \( G_t^{Rt} \in \mathbb{R}^{k \times d(d+1)/2} \). Final output of the layer consists of first- and second- order interactions, denoted by \( H_t^{Rt} = (F_t^{Rt} \mid G_t^{Rt}) \), which is then aggregated via sum pooling to generate an observation vector as $o_t^{Rt} = \sum_{l=1}^{d(d+3)/2} H_{t,l}^{Rt} \in \mathbb{R}^k$.

Moreover, given the importance of patient trajectory information for understanding the patient’s current context, a temporal contextual state vector \( c_t^{Rt} \). This vector captures the recent history of the patient's states by applying an exponential decay to the previous observation vectors, resulting in $c_t^{Rt} = \sum_{i=t-3}^{t-1} e^{-(t-i)} o_i^{Rt} \in \mathbb{R}^k$. 

The final core-state vector at the root level at time $t$ is constructed by concatenating the current observation vector \( o_{Rt} \) with the temporal contextual state vector \( c_{Rt} \), resulting in $s_t^{Rt} = (o_t^{Rt} \mid c_t^{Rt}) \in \mathbb{R}^{2k}$. This vector is learned end-to-end during the training of the root agent, enabling the model to capture immediate and historical physiological measurements effectively. The reduced dimensionality to $2k$ in the dense space also simplifies the complexity of subsequent decision-making processes. 

\noindent\textbf{\textit{Targeted State Representations:}} 
The generic latent embeddings $E^{Rt}$ learned at the root level are fine-tuned to generate targeted (organ-level) state representations that reflect unique characteristics and critical features relevant to each organ. Accordingly, 
% using a similar approach, 
we transform the raw features $x_t$ using specialized latent embeddings $E^{Neu}$, $E^{Car}$ and $E^{Ren}$, to obtain organ-specific state representations $s_t^{Neu}$, $s_t^{Car}$ and $s_t^{Ren}$. They are used as input states for training the corresponding sub-agents ($M_{Neu}$, $M_{Car}$ and $M_{Ren}$). A concatenated targeted state representation $s_t^{OMix} = \left[ s_t^{Neu} | s_t^{Car} | s_t^{Ren}\right]$ is used to train organ mixture agents ($M_{OMix}$, $M^{Neu}_{OMix}$, $M^{Car}_{OMix}$ and $M^{Ren}_{OMix}$), ensuring they are trained on a comprehensive representation of all organs.
\subsubsection{Actions}
% All agents operate within factored action spaces as introduced in by the proposed hierarchical task decomposition. Factored action spaces reduce the learning burden of each agent and prevents the spread of low-quality data to other subspaces. 
% In total, seven treatments were chosen, based on clinical expertise and data availability. Dosages for each treatment were discretized into five level based on quantiles. Two treatments were considered for the cardiovascular system: IV fluids (IV) and vasopressors (Vaso), three for neuro: anesthetics (S1), analgesics (S2), and sedatives (S3), and two for renal: diuretics and dialysis. Unlike the other treatments, dialysis is a binary action: active or inactive. As simultaneous usage of diuretics and dialysis occurs only in  very rare circumstances, and data for such a combination is lacking, we did not consider any mixing between these two renal treatments. As described, our solution flexibly handles organ systems with different degrees of complexity and configurations of actions. By utilizing `mixture' agents, our solution effectively manages mixing of multiple treatments from both within and between organ systems; these agents can learn consequential drug-drug interactions which may not always be explicitly considered in real clinical settings. 

In addition to the agent options discussed above for higher-level agents, this section provides detailed information on the actions of leaf-level agents. According to our hierarchical task decomposition approach, each agent operates within a factored action space. This helps reduce the learning burden on each agent and prevents the effects of potentially low-quality samples across subspaces. Based on clinical expertise and data availability, we selected six treatments. Two for each the Cardiovascular system: IV fluids (IV) and vasopressors (Vaso), Neuro system: anesthetics (S1) and analgesics (S2), and Renal system: diuretics and dialysis. These organ systems and their respective treatments were chosen by a physician according to clinical priority in intensive care patients. Each treatment plays a unique and critical function within its biological system and requires careful control to ensure efficacy and safety. Our framework is scalable to any number of organ systems and treatments. The proposed hierarchy is generic which allows integration of agents operating on continuous and discrete action spaces. However, we tested the approach using discrete action spaces. Dosages for each treatment were discretized into five levels (no-action + 4 quantiles), except dialysis, which is a binary action (active or inactive). Proposed hierarchy supports mixing of any treatment within or across systems, aligning with common clinical practices derived from our data analysis and clinical consultations. For example, we enforced exclusive use of one renal treatment at a time—either diuretics or dialysis, and allowed all other treatment combinations. This showcases the hierarchy's adaptability to incorporate such domain specific constrains. Accordingly, our solution flexibly handles organ systems with different degrees of complexity and configurations of actions, enabling the effectiveness of the solution in real, complex clinical settings.

\subsubsection{Reward Function}
% We used a mortality based terminal reward, where trajectories leading to patient ,whereassurvival or death received a large positive \( +R \) or negative \( -R \) reward, respectively. However, using only a terminal reward for long sequences can lead to exploration issues due to reward sparsity. Therefore, additionally we introduce a clinically guided intermittent reward function that rewards or penalises each action based on immediate health outcomes ({\color{pink}see Appendix, Section \textit{RL Action}}).
We used a hybrid reward system, combining a mortality-based terminal reward—positive ($+R$) for survival, and negative ($-R$) for death—with clinically guided intermittent rewards to adjust rewards based on immediate health outcomes. Specifically, we define intermediate rewards based on changes in the Sequential Organ Failure Assessment (SOFA) score and blood lactate levels—two validated surrogates of physiological deterioration. The reward penalizes both persistently high SOFA scores and increases in SOFA or lactate between successive time steps, while assigning positive reward for improvements. The overall form is:

\begin{equation}
R(s_t, s_{t+1}) = 
\begin{cases}
+R, & \text{if } s_{t+1} \in \mathcal{S}_T \cap \mathcal{S}_{sur} \\
-R, & \text{if } s_{t+1} \in \mathcal{S}_T \cap \mathcal{S}_{dec} \\
R_{im}(s_t, s_{t+1}), & \text{if } s_{t+1} \notin \mathcal{S}_T
\end{cases}
\label{eq:reward_total}
\end{equation}

\vspace{1em}

\begin{equation}
\begin{split}
R_{im}(s_t, s_{t+1}) = & \; C_0 \big( s_{t+1}^{SOFA} = s_t^{SOFA} \; \land \; s_{t+1}^{SOFA} > 0 \big) \\
& + C_1 \big( s_{t+1}^{SOFA} - s_t^{SOFA} \big) \\
& + C_2 \tanh \big( s_{t+1}^{lactate} - s_t^{lactate} \big)
\end{split}
\label{eq:reward_intermediate}
\end{equation}

where $C_0 = -0.025$, $C_1 = -0.125$, and $C_2 = -2$. The $\tanh$ term ensures that large lactate swings do not disproportionately affect learning.

The chosen approach addresses the challenges of credit assignment and reward sparsity in long sequences. They ensure that immediate health improvements are reflected in decision-making while supporting adaptable, cross-organ policy learning applicable across diverse healthcare settings. Our reward design follows established structures in healthcare RL research; however our framework itself is generalizable as it is not tied to an specific reward function. Where terminal rewards suffice, our framework can operate without intermittent rewards.

\subsection{Q Learning}

Traditional RL frameworks, using Markov Decision Processes (MDPs), often assume actions are executed instantaneously and have uniform durations. This simplifies modeling, but fails to capture real-world complexities with actions spanning multiple steps and varying duration. 

Therefore we utilize an integration of an options framework with semi-MDPs and decentralized MDPs within a structured hierarchical system. This approach enables an agent to invoke options that ranges from primitive (one-step) to extended (multiple-step) actions at any point. Then the agent regains control to initiate another option only after the previous one completes. Accordingly, the traditional action-value function is adapted to an option-value function $Q^{\pi}(s,o)$ which indicates the value of invoking option $o$ in state $s$ under a policy $\pi$. This is defined as the expected sum of discounted future rewards: 
\begin{equation*}
% \label{eq:q_pi}
Q^{\pi}(s,o) = \mathbb{E}\{r_{t+1}+\gamma r_{t+2}+\gamma^2 r_{t+3}+\dots\mid\epsilon^{\pi}(o,s,t) \}
\end{equation*}
where $\gamma$ is a discount factor and $\epsilon^{\pi}(o,s,t)$ is the execution of option $o$ in state $s$ from time $t$ to its termination. Unlike the standard options framework, our solution contains independent and cooperative agents controlling options (see Section \ref{sec:hd}). Thus, root level Q-values are updated as follows:
\begin{equation}
\label{eq:q}
Q(s,o) \leftarrow Q(s,o) + \alpha \left[ Q_{tgt}(s,o) - Q(s,o) \right]
\end{equation}
where $Q_{tgt}(s,o)$ is determined based on the controlling agent and the option $o$ as follows:
\begin{equation*}
% \label{eq:q_tgt}
\begin{aligned}
Q_{tgt}(s,o) = & \begin{cases} 
r, & \text{if } o = A_{Rt}^{0} \\
\max\limits_{a \in A_{Neu}} Q_{Neu}(s,a), & \text{if } o = A_{Rt}^{Neu} \\
\max\limits_{a \in A_{Car}} Q_{Car}(s,a), & \text{if } o = A_{Rt}^{Car} \\
\max\limits_{a \in A_{Ren}} Q_{Ren}(s,a), & \text{if } o = A_{Rt}^{Ren} \\
\max\limits_{a \in A_{OMix}} Q_{OMix}(s,a), & \text{if } o = A_{Rt}^{OMix}
\end{cases}
\end{aligned}
\end{equation*}

where $r$ is the immediate reward for selecting no-action $A_{Rt}^{0}$, $A_{Neu}$, $A_{Car}$, $A_{Ren}$ and $A_{Rt}^{OMix}$ are the action spaces for Neuro, Cardio, Renal agents, and their combined action space $A_{Neu}\times A_{Car}\times A_{Ren}$. Value functions \( Q_{Neu}(s,a) \), \( Q_{Car}(s,a) \), and \( Q_{Ren}(s,a) \) for corresponding agents are updated using Temporal Difference (TD) learning as follows:

\begin{equation}
\label{eq:q_oran_level}
\begin{aligned}
Q_{Neu}(s, a) \leftarrow &Q_{Neu}(s, a) + \alpha \left[ y^{Neu} - Q_{Neu}(s, a) \right]\\
Q_{Car}(s, a) \leftarrow &Q_{Car}(s, a) + \alpha \left[ y^{Car} - Q_{Car}(s, a) \right]\\
Q_{Ren}(s, a) \leftarrow &Q_{Ren}(s, a) + \alpha \left[ y^{Ren} - Q_{Ren}(s, a) \right]\\
\end{aligned}
\end{equation}

where \( y^{X} = r + \gamma \max_{a'} Q_{X}(s', a'; \theta^{-}) \) for \( X \in \{Neu, Car, Ren\} \) are TD targets computed from target networks with parameters \( \theta^{-} \). The Q values for each system are updated using observed rewards and estimated values of the next state (\( s'\)).

$M_{OMix}$ is trained using the QMix architecture \cite{rashid2020monotonic}, combining individual value functions of its sub-agents ($M_{Omix}^{Neu}$,$M_{Omix}^{Car}$, and $M_{Omix}^{Ren}$) into a unified value function ($Q_{OMix}$). QMix supports cooperative training and consistent decision-making across agents by enforcing a monotonicity constraint. The constraint keeps the weights of the mixing network non-negative, ensuring that the $argmax$ operation on $Q_{OMix}$ is consistent with those from the sub-agents. Hence, $Q_{OMix}$ is represented as follows:

\begin{equation}
\label{eq:argmax}
\begin{aligned}
& \mathrm{arg\,max}_{a \in A_{OMix}} Q_{OMix}(s,a) = \\
& \left(
\begin{aligned}
&\mathrm{arg\,max}_{a_{OMix}^{Neu} \in A_{OMix}^{Neu}} Q_{OMix}^{Neu}(s_{OMix}^{Neu}, a_{OMix}^{Neu}) \\
&\mathrm{arg\,max}_{a_{OMix}^{Car} \in A_{OMix}^{Car}} Q_{OMix}^{Car}(s_{OMix}^{Car}, a_{OMix}^{Car}) \\
&\mathrm{arg\,max}_{a_{OMix}^{Ren} \in A_{OMix}^{Ren}} Q_{OMix}^{Ren}(s_{OMix}^{Ren}, a_{OMix}^{Ren})
\end{aligned}
\right)
\end{aligned}
\end{equation}

where $s_{OMix}^{Neu} = \left[ s_t^{OMix}, a_{Car}, a_{Ren} \right]$ concatenates the targeted state representation and communicated dosage outputs ($a_{Car}$ and $a_{Ren}$) from the opposite organ-specific agents ($M_{Car}$ and $M_{Ren}$) as described under \textit{Hierarchical Decomposition}. $s_{OMix}^{Car}$ and $s_{OMix}^{Ren}$ are formed analogously.

\subsubsection{Training Process:}

We used a two phase approach to train the proposed hierarchical model (see \textit{Algorithm 1} in Appendix). In phase one, the root agent ($M_{Rt}$) is trained first. Resulting latent embeddings $E^{Rt}$ are then fine-tuned during the subsequent training of organ-specific agents ($M_{Neu}$, $M_{Car}$, and $M_{Ren}$). In the process, targeted embeddings ($E^{Neu}$, $E^{Car}$ and $E^{Ren}$) are learned and formulate \textit{Targeted State Representations}. These representations are used to train the corresponding lower-level sub-agents. Independent agents directly learn $Q$ values from received rewards (see Equation \ref{eq:q_oran_level}), whereas cooperative agents $M_{Omix}^{Neu}$,$M_{Omix}^{Car}$, $M_{Omix}^{Ren}$ are trained using shared rewards (Equation \ref{eq:argmax}).  
In phase two, we integrate the full hierarchy by using the trained organ-specific agents from phase one. $Q(s,a)$ values from these agents are used to retrain $M_{Rt}$ (see Equation \ref{eq:q}), and no other agent is retrained. After training, at each timestep, $M_{Rt}$ evaluates the patient state $s_{t}^{Rt}$ and select either \textit{no-action}, or invokes lower level agents for final treatment and dosage recommendations.

\section{Experiments}
\subsection{Datasets}
% We collect data for 30,440 patients who fulfilled the Sepsis-3 crtieria from the Multi-parameter Intelligent Monitoring in Intensive Care (MIMIC-IV) database \cite{johnson2020mimic} (Table \ref{tab:data_summary}). The data is filtered from 24 hours before sepsis diagnosis and up to 48 hours after diagnosis forming a maximum 72 hour window per patient barring death or ICU discharge. Patient state data collected includes 48 physiological measurements including vital signs, laboratory test results, severity scores, demographics, etc ({\color{pink}Section X}). State and actions data were aggregated into four-hourly windows to generate uniform patient data sequences. For each four hour window, we calculated each treatment's total administered dosage by multiplying its infusion rate with the overlapping duration. We subsequently added any additional volume of drug administered via IV push within the time window to acquire the final four-hourly dosage. Treatments which consisted of more than one drug i.e. vasopressors, S2, and diuretics were converted into standard—norepinepherine \cite{kotani2023updated}, fentanyl \cite{mcpherson2010demystifying}, and furosemide \cite{konerman2022michigan} equivalents respectively. The data was split into 75\% training and 25\% testing sets. 

We collected data on 30,440 patients under Sepsis-3 criteria from the popular MIMIC-IV database \cite{johnson2020mimic}. The data spans from 24 hours pre-diagnosis to 48 hours post-diagnosis forming a maximum 72 hour window per patient, barring death or ICU discharge. We defined a continuous state-space which included 44 physiological measurements including vital signs, laboratory test results, severity scores and demographics. State and actions data were aggregated into four-hourly windows to generate uniform patient data sequences. The patient trajectories were randomly split into 75\% training and 25\% testing sets. Additionally, we collected data for 4,390 sepsis patients from the AmsterdamUMC dataset \cite{thoral2021sharing} to conduct an external validation of our trained models using the same set of state and action variables. High level statistics of the cohorts are listed in Table \ref{table:cohort_characteristics}. See Appendix, Section \textit{RL Actions} and \textit{RL States} for more information on the state and action spaces.

\begin{table}[H]
\centering
\footnotesize
\setlength{\tabcolsep}{8pt}  % Adjusted to fit better while keeping spacing readable
\caption{Cohort characteristics for MIMIC-IV and AmsterdamUMC}
\begin{tabular}{@{}p{2cm}@{}p{1.5cm}@{}p{1.2cm}@{}p{1.2cm}@{}p{1.5cm}@{}p{1.5cm}@{}}
\toprule
\textbf{Dataset} & \textbf{Cohort} & \textbf{\% Female} & \textbf{Age (Mean)} & \shortstack{\textbf{ICU LOS} \\ \textbf{(Mean)}} & \textbf{Population} \\
\midrule
MIMIC-IV
  & Overall        & 42.11 & 66.14 & 5d 4h   & 30,440 \\
  & Non-Survivors  & 44.00 & 69.77 & 6d 12h  & 4,948  \\
  & Survivors      & 41.75 & 65.44 & 4d 22h  & 25,492 \\
\midrule
AmsterdamUMC
  & Overall        & 38.82 & 61.71 & 7d 8h   & 4,390  \\
  & Non-Survivors  & 39.49 & 66.64 & 9d 9h   & 961    \\
  & Survivors      & 38.63 & 60.33 & 6d 18h  & 3,429  \\
\bottomrule
\end{tabular}
\label{table:cohort_characteristics}
\end{table}

% \begin{table}[H]
% \centering
% \footnotesize  % Smallest readable font
% % \renewcommand{\arraystretch}{0.95}  % Reduce row height
% \setlength{\tabcolsep}{25pt}         % Reduce column padding
% \caption{Cohort characteristics for MIMIC-IV and AmsterdamUMC}
% \begin{tabular}{@{}p{2cm}@{}p{1cm}@{}p{1.0cm}@{}p{1.0cm}@{}p{1cm}@{}p{1.5cm}@{}}
% \toprule
% \textbf{Dataset} & \textbf{Cohort} & \textbf{\% Female} & \textbf{Age (Mean)} & \textbf{ICU LOS (Mean)} & \textbf{Population} \\ 
% \midrule
% {MIMIC-IV}
%   & Overall        & 42.11 & 66.14 & 5d 4h   & 30,440 \\
%   & Non-Survivors  & 44.00 & 69.77 & 6d 12h  & 4,948  \\
%   & Survivors      & 41.75 & 65.44 & 4d 22h  & 25,492 \\
% \midrule
% {AmsterdamUMC}
%   & Overall        & 38.82 & 61.71 & 7d 8h   & 4,390  \\
%   & Non-Survivors  & 39.49 & 66.64 & 9d 9h   & 961    \\
%   & Survivors      & 38.63 & 60.33 & 6d 18h  & 3,429  \\
% \bottomrule
% \end{tabular}
% \label{table:cohort_characteristics}
% \end{table}

% \subsection{Model Parameters}
% {\color{pink} Can move to Appendix to save space.}

\subsection{Baselines}
We evaluated our model performance against multiple baselines, including single- (D3QN-S and SoftAC-S) and multi-agent systems under independent (D3QN-O) and cooperative (QMix-O and QMix-T) learning approaches:
\setlist{nolistsep}
\begin{itemize}[noitemsep]
    \item \textbf{Clinician}: Policy derived from clinician's recorded action trajectories in the test set.
    \item \textbf{Single D3QN (D3QN-S)} \cite{raghu2017deep}: Single-agent D3QN predicting all treatment and dosage combinations in a flattened action space. We used the original implementation with tuned hyperparameters.
    \item \textbf{Single SoftAC (SoftAC-S)} \cite{haarnoja2018soft}: A single-agent Soft Actor-Critic predicting all action combinations in a flattened action space. We used the original implementation and tuned hyperparameters.
    \item \textbf{Organ-specific D3QN (D3QN-O)}: Three independent D3QN agents for Neuro, Cardio and Renal systems. Models are trained using all available samples, including those treated exclusively for the target organ and those mixed with other organ treatments. Averaged quantitative metrics are reported.
    \item \textbf{Treatment-specific D3QN (D3QN-T)}: Five independent D3QN agents for S1, S2, IV, vaso, and diuretics/dialysis. Models are trained using all available samples, including those using single-treatments and mixed-treatments. Averaged quantitative metrics are reported.
    \item \textbf{Organ-coordinated QMix (QMix-O)}: Trained end-to-end with three cooperative agents, each corresponding to an organ system operating exclusively within its factored action spaces. Predicts treatments across organs by cooperation, using a QMix mixing network. 
    \item \textbf{Treatment-coordinated QMix (QMix-T)}: Uses five cooperative treatment-level agents (S1, S2, IV fluids, vasopressors, diuretics/dialysis), learning cooperatively via a QMix network. 
\end{itemize}

All models used the same state representations, action discretization methods, and reward functions. All codes for data processing and model training are included in the supplementary materials and will publicly available upon acceptance.

\subsection{Results and Discussion}
We present and discuss our experimental results as answers to four key research questions as follows.

% \begin{figure}
%     \centering
%     \includegraphics[width=1\linewidth]{figures/qvsmort_all_.pdf}
%     \caption{Mortality vs. expected return for all models. The shaded area represents standard errors.}
%     \label{fig:q_vs_mort}
% \end{figure}

% \begin{figure}
%     \centering
%     % \vspace{0.5cm}
%     \includegraphics[width=1\linewidth]{figures/vplot3.png}
%     \caption{Difference in clinician and model's learnt policy dosages (x-axis) and mortality (y-axis), for the best-performing single-agent baseline (SoftAC), best-performing multi-agent baseline (QMix-T), and proposed model.}
%     \label{fig:mortality_vs_dosagediff}
% \end{figure}

\begin{table}%[ht]
\small
\centering
\caption{Policy Evaluation on MIMIC-IV and AmsterdamUMC}
\label{table:quant_eval}
\begin{tabular}{lcc|cc}
\toprule
\textbf{Policy} & \multicolumn{2}{c|}{\textbf{MIMIC-IV}} & \multicolumn{2}{c}{\textbf{External Validation}} \\
                & \textbf{V} & \textbf{Mortality (\%)} & \textbf{V} & \textbf{Mortality (\%)} \\
\midrule
Clinician       & 9.08   & 16.27             & 7.74   & 21.89 \\
SoftAC-S          & -3.26  & 20.52 $\pm$ 0.68  & -6.22   & 28.59 $\pm$ 0.85 \\
D3QN-S        & -1.29  & 19.79 $\pm$ 0.77  & -6.87   & 29.13 $\pm$ 0.92\\
D3QN-O          & 12.74   & 14.57 $\pm$ 0.30  & 8.01   & 20.57 $\pm$ 0.50\\
D3QN-T          & 13.05   & 13.62 $\pm$ 0.43  & 10.24   & 19.91 $\pm$ 0.45\\
QMIX-O          & 15.45   & 13.29 $\pm$ 0.37  & 8.70   & 20.12 $\pm$ 0.55\\
QMIX-T          & 17.16   & 12.74 $\pm$ 0.30  & 11.07   & 18.56 $\pm$ 0.46\\
\textbf{Proposed} & \textbf{21.73} & \textbf{9.95 $\pm$ 0.33} & \textbf{12.95} & \textbf{17.34 $\pm$ 0.43} \\
\bottomrule
\end{tabular}
\end{table}

\begin{figure}[t]
    \centering
    % First row
    \begin{subfigure}{.28\linewidth}
      \centering
      \includegraphics[width=\linewidth]{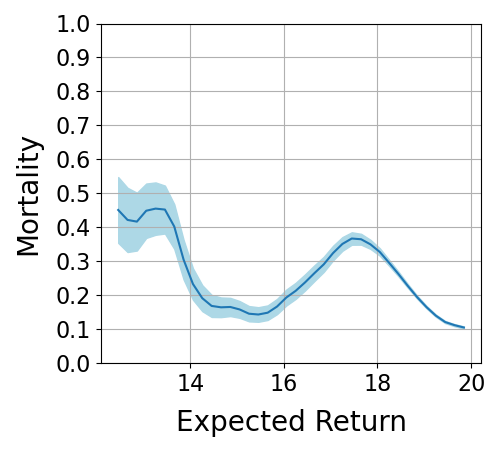}
      \caption{\tiny SoftAC-S}
    \end{subfigure}
    \hspace{0.04\linewidth}
    \begin{subfigure}{.28\linewidth}
      \centering
      \includegraphics[width=\linewidth]{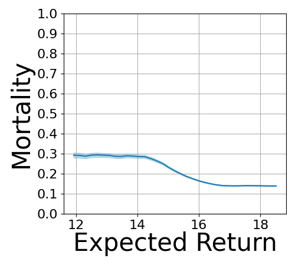}
      \caption{\tiny D3QN-S}
    \end{subfigure}
    \hspace{0.04\linewidth}
    \begin{subfigure}{.28\linewidth}
      \centering
      \includegraphics[width=\linewidth]{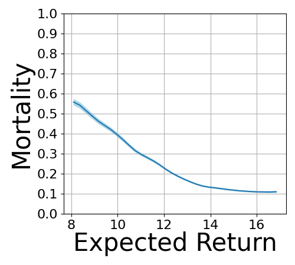}
      \caption{\tiny QMix-O}
    \end{subfigure}

    \vspace{0.4cm} % vertical space between rows

    % Second row
    \begin{subfigure}{.28\linewidth}
      \centering
      \includegraphics[width=\linewidth]{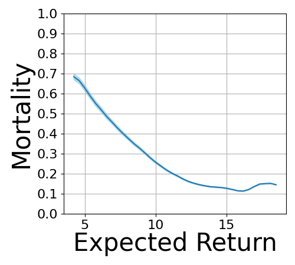}
      \caption{\tiny QMix-T}
    \end{subfigure}
    \hspace{0.04\linewidth}
    \begin{subfigure}{.28\linewidth}
      \centering
      \includegraphics[width=\linewidth]{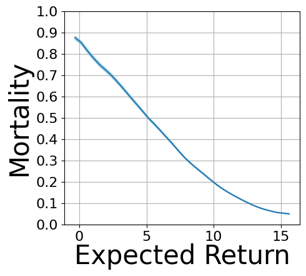}
      \caption{\tiny Proposed}
    \end{subfigure}
    \hspace{0.04\linewidth}
    \begin{subfigure}{.28\linewidth}
      \centering
      \includegraphics[width=\linewidth]{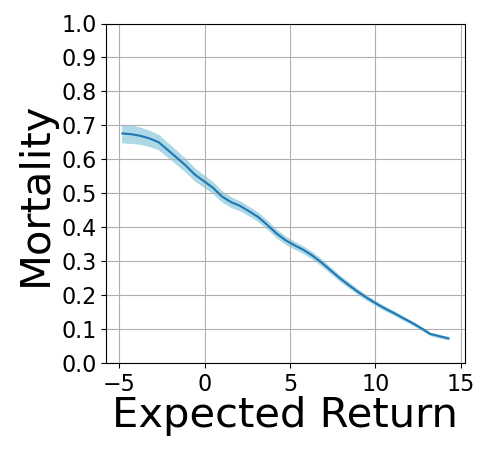}
      \caption{\tiny Prop. (External Val.)}
    \end{subfigure}

    \caption{Mortality vs. expected return for all models. The shaded area represents standard errors.}
    \label{fig:q_vs_mort}
\end{figure}
\subsubsection{\textbf{RQ1: Does the proposed solution effectively learns a superior treatment policy?}}
We evaluate the performance of learned policies using off-policy quantitative and qualitative metrics (see Table \ref{table:quant_eval}). They do not require a simulator, but adopt a well-established approach using offline learning on retrospective data. 
% supported by QMIX’s off-policy training. 
These metrics, widely used in healthcare RL enable safe, effective learning from clinical outcomes without real-time exploration risks \cite{raghu2017deep,komorowski2018artificial}.
% \begin{table}[h]
% \centering
% \small
% \caption{Comparison of off-policy evaluation metrics}
% \label{tab:eval_metrics}
% \begin{tabular}{|c|c|c|c|}
% \hline
% \textbf{Model} & \textbf{ESS} & \textbf{V} & \textbf{Mortality (\%)} \\ \hline
% Clinician & 89115 & 18.98 & 16.27 \\ \hline
% D3QN-S & 12 & -1.21 & 18.69 ± 0.45 \\ \hline
% SoftAC-S & 975 & -0.19 & 18.52 ± 0.57 \\ \hline
% D3QN-O & 99 & 0.37 & 14.47 ± 0.34 \\ \hline
% D3QN-T & 373 & 19.42 & 14.02 ± 0.41\\ \hline
% QMix-O & 41 & 13.80 & 13.84 ± 0.15 \\ \hline
% QMix-T & 602 & 25.83 & 11.29 ± 0.28 \\ \hline
% \textbf{Proposed} & \textbf{7233} & \textbf{30.04} & \textbf{8.81 ± 0.24} \\ \hline
% \end{tabular}
% \end{table}
\begin{figure*}[t]
    \centering
    \includegraphics[width=0.7\linewidth]{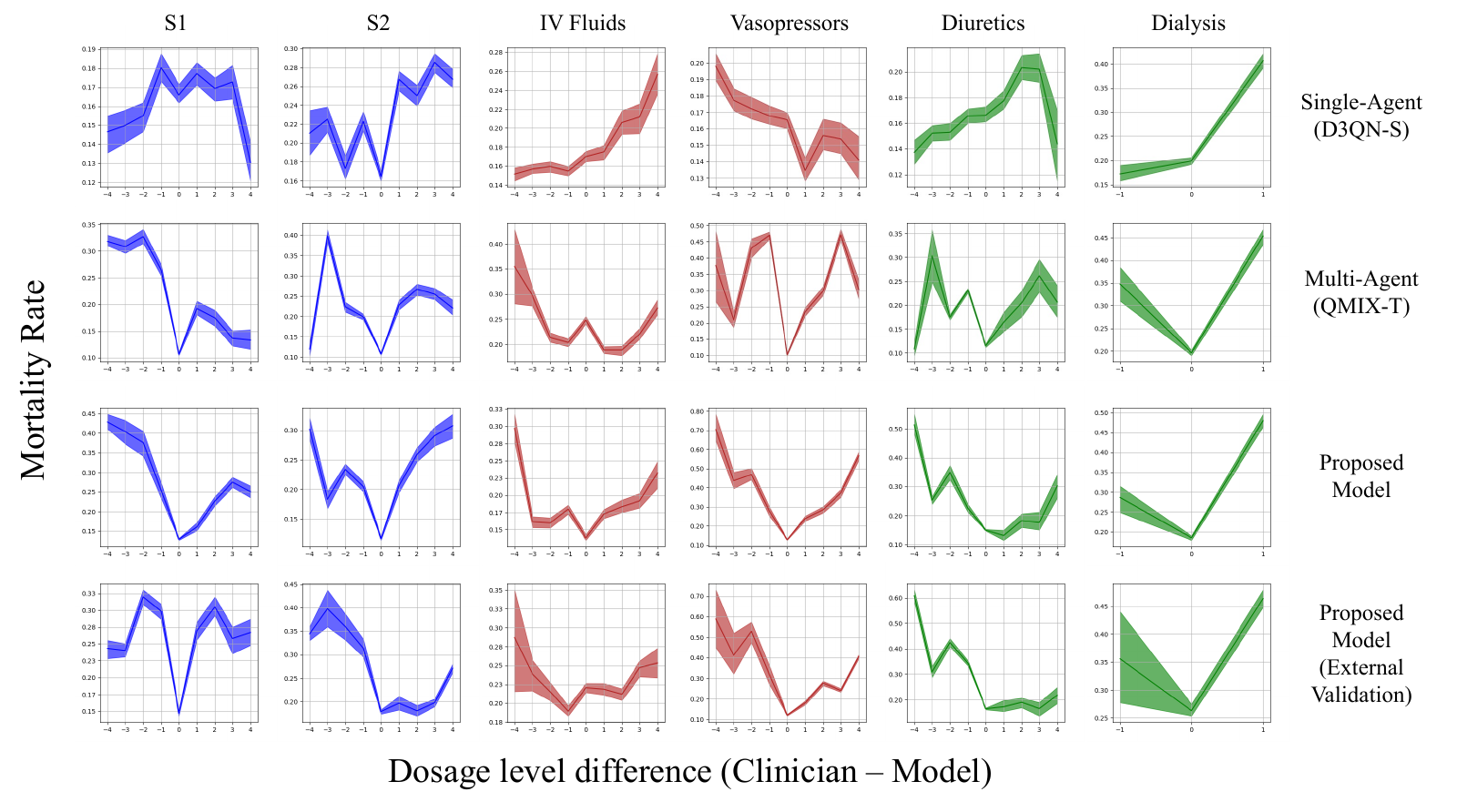}
    \caption{Dosage differences (x-axis) versus mortality (y-axis) for the top single-agent baseline SoftAC, multi-agent baseline QMix-T, proposed model on the testing set, and followed lastly by the proposed model on the external validation set, with colors blue, red, and green denoting different organ systems.}
    \label{fig:mortality_vs_dosagediff}
\end{figure*}

\noindent\textit{1) Average Returns:} 
% The performance of learned policies is quantitatively evaluated using their estimated average returns. 
We use Consistent Weighted Per-Decision Importance Sampling (CWPDIS), a standard OPE method to estimate expected returns from each policy \(V^{CWPDIS}\) ($V$). \cite{thomas2015safe}. A higher \(V\) corresponds to a more effective policy.  Both single agent baselines (D3QN-S and SoftAC-S) result in negative average returns. This indicates the single agent systems' inability to learn effective policies in this complex setting. Meanwhile, the multi-agent model QMix-T was the best performing baseline according to OPE returns. This suggests that for our particular multi-organ problem, MARL approaches with some form of cross-agent communication can be beneficial for policy learning.
Overall, our model considerably outperformed all baselines and the clinician policy, obtaining the highest estimated returns ($V=21.73$). This suggests that the model learnt a superior policy which acts in accordance to high-return clinician actions while deviating from low return ones. Results of the external validation show consistent, albeit more modest results. The smaller improvements in the OPE returns on the AmsterdamUMC dataset may be explained by differences in patient populations (U.S vs European cohorts) and the differences in clinical setting/protocols, among other reasons. 
 
\noindent\textit{2) Mortality Rate:} In line with literature, we estimate mortality rates using the clinician policy's relationship between mortality and expected returns. We do this by categorizing expected returns from patient trajectories into bins and calculate the mortality rate for each bin. The resulting relationship between the expected returns and mortality is used to estimate mortality rates for learned policies based on their expected returns. The single-agent baselines (SoftAC-S and D3QN-S) showed approximate 26\% and 22\% increases in mean mortality rate compared to the clinician respectively, indicating a failure to learn an improved policy. In contrast, cooperative multi-agent baselines (QMix-O and QMix-T) showed a decrease of 12.2\% and 21.7\% in mortality rate. The non-cooperative multi-agent baselines (D3QN-O and D3QN-T) outperformed single agent baselines, but were less effective than the cooperative multi-agent baselines, highlighting the importance of collaboration among agents to obtain superior policies. Our model had the highest decrease in mean mortality by 38.9\%, highlighting its ability to learn a policy that could considerably improve patient survival compared to state-of-the-art models and clinician policy. While we see a relatively smaller improvement in mortality rate in our external validation (20.9\% decrease), our proposed model still achieves the lowest estimated mortality rate among all baselines, supporting the performance advantages of our approach.

\noindent\textit{3) Mortality vs. Expected Return:} We further evaluate the efficacy of the learned policies by analyzing the correlation between mortality and expected returns (see Figure \ref{fig:q_vs_mort}). An effective policy should display a strong negative correlation, where higher expected returns translate to lower mortality, and low expected returns to higher mortality. This relationship reflects the policy's ability to discern optimal actions from sub-optimal actions.  The multi-agent baselines (QMix-O and QMix-T) display steeper negative curves than the single-agent baselines (D3QN-S and SoftAC-S). However, overall we see that none of the baselines consistently maintained the negative correlation throughout the plot—with plateauing and positive correlations at some points in the plot. In contrast, the proposed model showed the steepest negative correlation with a consistent negative relationship, indicating the highest ability to improve patient survival. In our external validation, we see the same consistent negative relationship, indicating a good degree of generalizability.

\noindent\textit{4) Mortality vs. Difference in Recommended Dosage:} For each intervention, we analyzed the relationship between mortality rates, and the recommended dosage differences between clinician-administered and learned policies. We estimated this relationship by categorizing quantile-level dosage differences into bins and computing mortality rates of each bin. An effective policy should align with clinician dosages that resulted in low mortality (x-axis=0), and increasingly deviate from those associate with increasing mortality, ideally forming a 'U' or 'V'-shaped curve centered at 0.

We showcase the performance of our model against the best-performing single-agent (D3QN-S) and multi-agent (QMix-T) models (see Figure \ref{fig:mortality_vs_dosagediff}). See Appendix, Section \textit{Experimental Results} for comparison across all baseline models. The single-agent model failed to achieve the desired V-shape for any treatment, indicating a failure to properly learn from the large action space. The multi-agent baseline shows comparatively better performance, again demonstrating the advantage of cooperative agents operating under factored action spaces. On the other hand, our proposed model most closely approximated the desired V-shaped relationship across treatments; these qualitative results were also  adequately replicated in our external validation, highlighting decent generalizability of our trained model.

% B. \textit{RQ2: Is the policy learnt by the our proposed solution superior to that of independent single-organ and single-treatment agents?}
% \newline
% The two baselines D3QN-O and D3QN-T are multi-agent models in which agents are both independently trained and executed. D3QN-O comprises of three organ-level agents, and D3QN-T, six treatment-level agents. While we see that these models can achieve lower mean mortality rate than the clinician's policy, they are outperformed by the cooperative multi-agent baselines (QMix-O and QMix-T) as well as our proposed model according to CWPDIS and mean mortality rate. This indicates that some form of cooperation among agents is advantageous for the complex, multi-organ treatment recommendation task  
\begin{figure}%[!t]
    \centering
    \includegraphics[width=\linewidth]{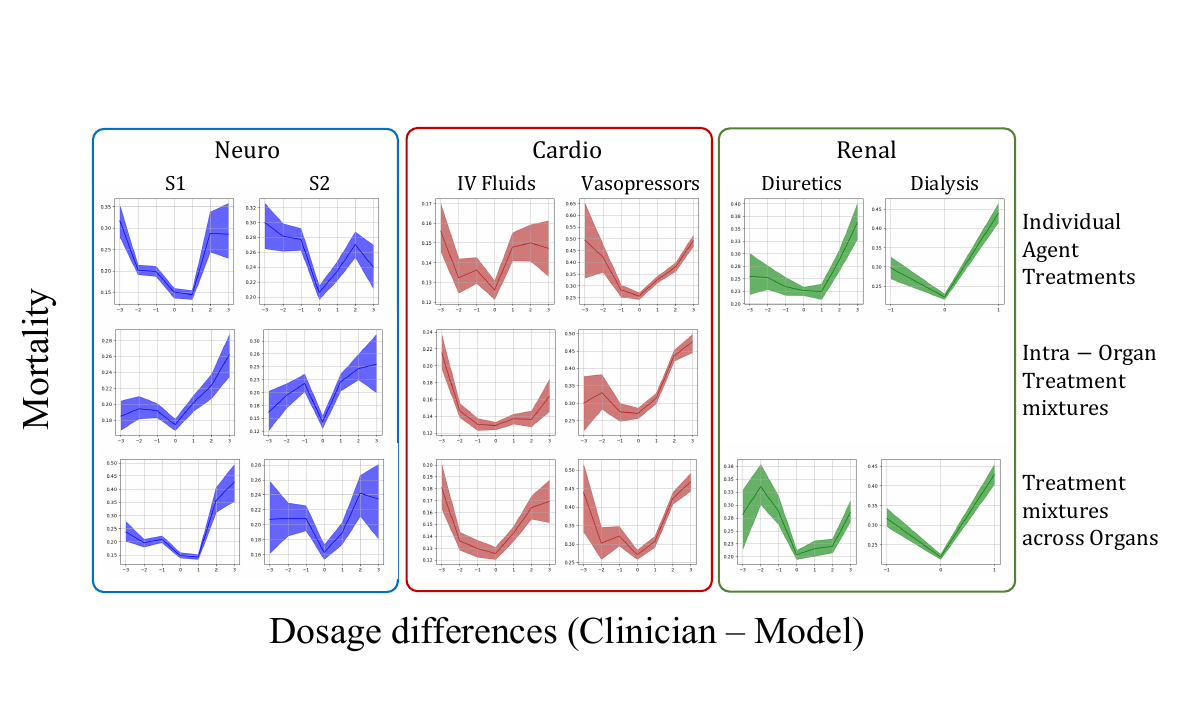}
    % \caption{Performances of individual agents in the hierarchy. Dosage differences (x-axis) versus mortality (y-axis) for individual agents in the hierarchy, with colors blue, red, and green denoting different organ systems.}
    \caption{Mortality versus dosage difference (Clinician – Model) for individual agents across Neuro (blue), Cardio (red), and Renal (green) systems. Rows represent: individual treatments (top), intra-organ mixtures (middle), and inter-organ coordination (bottom). Renal agents do not include intra-organ mixtures due to clinical exclusivity between diuretics and dialysis.}
    \label{fig:individual_agents}
\end{figure}

% \begin{figure}[!t]
%     \centering
%     \includegraphics[width=\linewidth]{figures/dummy_policycomp.png}
%     \caption{\tiny Comparison for survived trajectories.}
%     \label{fig:mortality_vs_dosagediff}
% \end{figure}

\subsubsection{\textbf{RQ2: Do individual agents learn effective local policies?}}
To assess if the agents operating within localized state-action subspaces learn effective treatment policies, We analyzed the relationship between mortality rates and discrepancies between the clinician and each agent recommended dosages (see Figure \ref{fig:individual_agents}). Across all three organ systems, we observe that most agents showcase the desired V shape. For example, the Vasopressors and S1 agents show steep increases in mortality when deviating from clinician dosages, showing the sensitivity of the interventions. Proposed model tend to match effective dosing and adaptively diverge in high-risk scenarios, which supports is clinical relevance.
Importantly, the renal sub-agents only appear in individual and cross-organ coordination rows as they do not contribute to intra-organ mixtures. This was intentionally modeled as renal treatments were found mutually exclusive in clinical practice, where diuretics and dialysis are found to be applied at different severity levels and rarely administered together. Enforcing exclusivity improves training stability and better reflects real-world treatment logic and flexibility in the proposed framework.

% In addition to evaluating the overall effectiveness of the learned hierarchical policy, we evaluated the independent learning capability of sub-agents that learn local policies within the hierarchy. Each agent operates in a factored state and action spaces targeting specific organs and interventions which enable focused learning with minimum cross-agent interferences. We analyzed the relationship between mortality rates and discrepancies between the clinician and each agent recommended dosages (see Figure \ref{fig:individual_agents}). All individual agents' policies closely followed the desired 'V' shape, indicating effective local policies across the hierarchy. This increases our confidence in the combined global policy and enhances the reliability of our proposed hierarchical task decomposition solution. 
\subsubsection{\textbf{RQ3: How does the learned policy compare to the clinician policy in the multi-organ treatment decision context?}} 
\begin{figure}[!t]
    \centering
    \begin{subfigure}{.48\linewidth}
        \includegraphics[width=\linewidth]{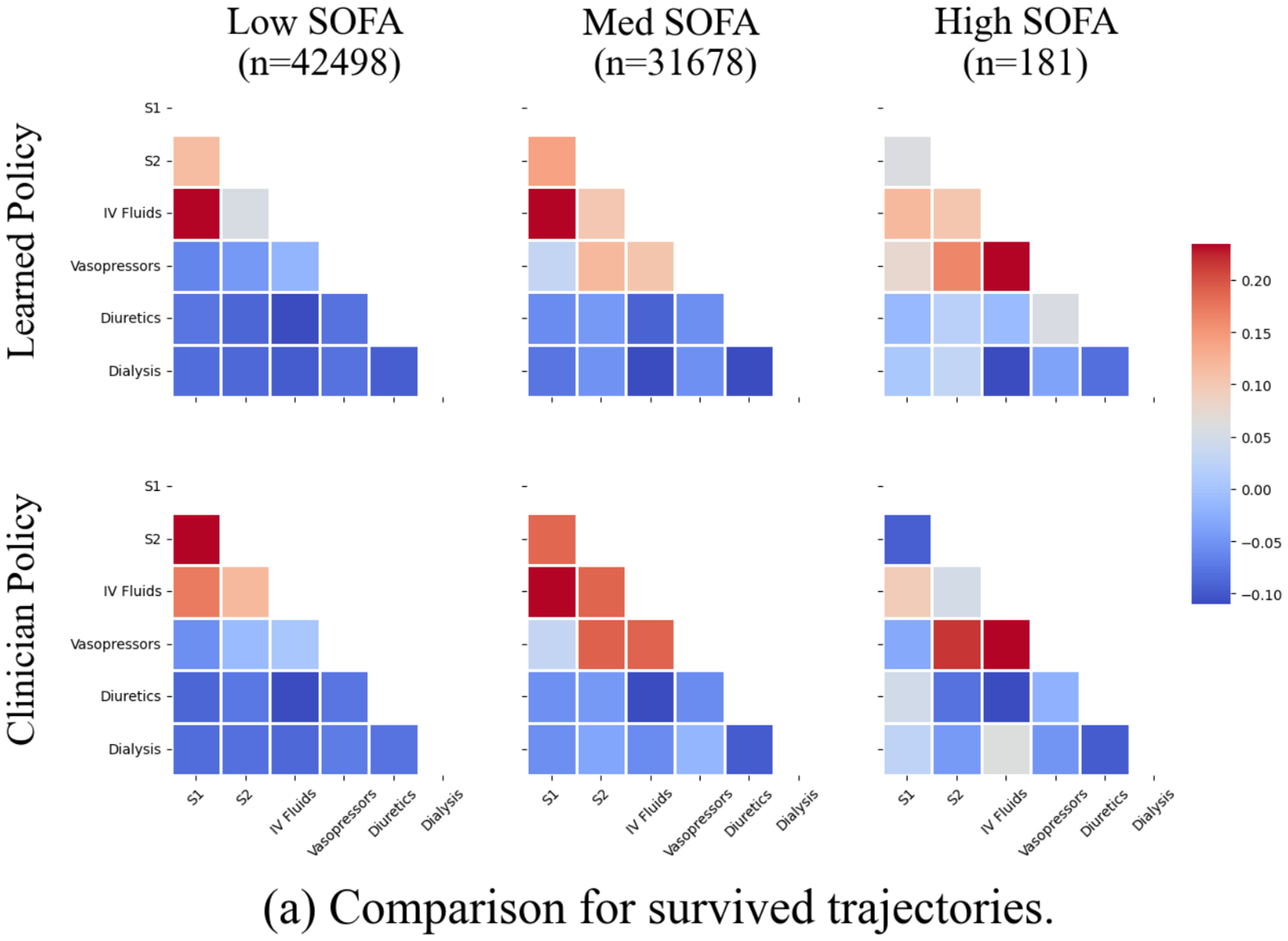}
        % \caption{\tiny Comparison for survived trajectories.}
    \end{subfigure}
    \begin{subfigure}{.48\linewidth}
        \includegraphics[width=\linewidth]{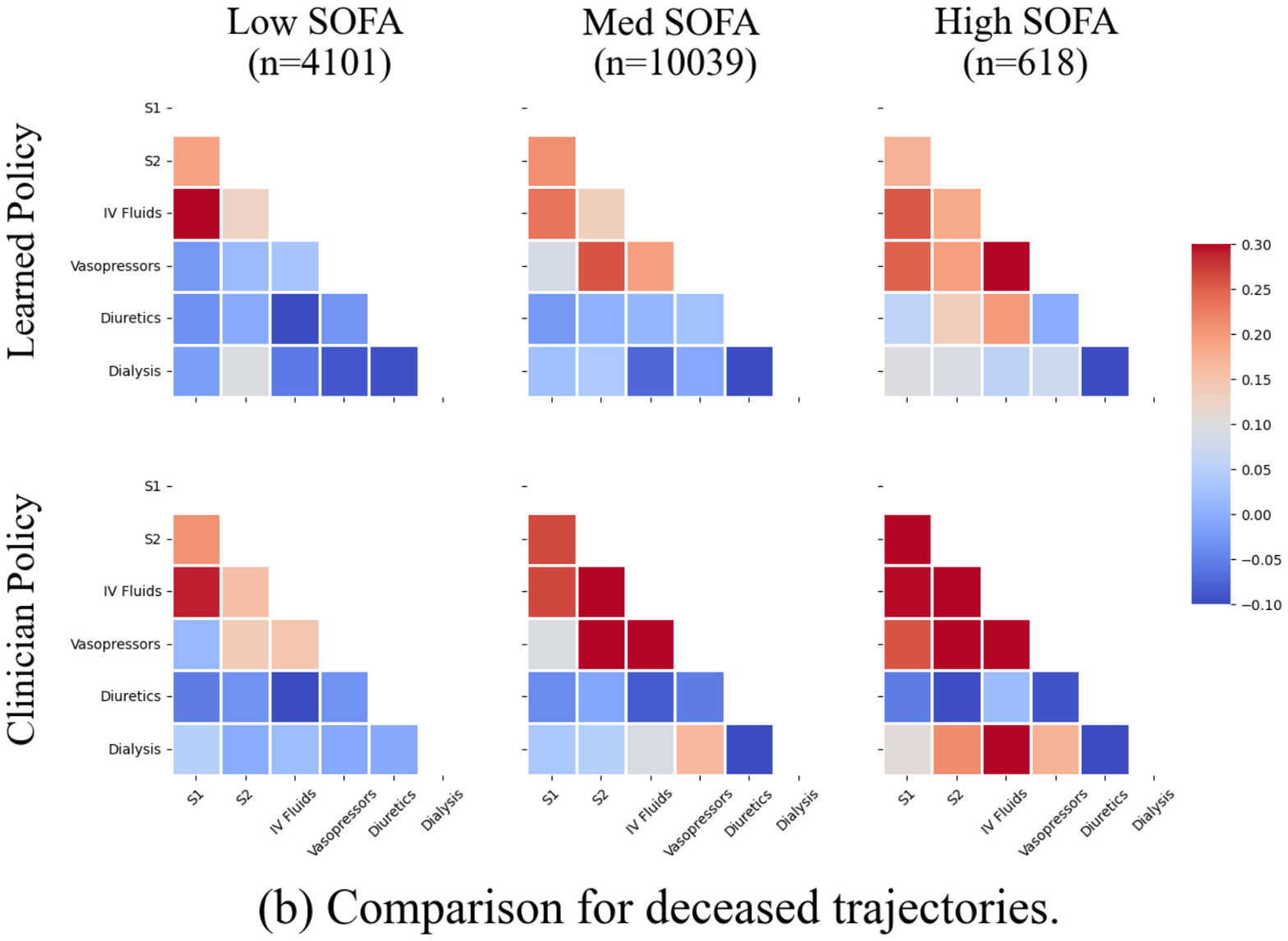}
        % \caption{\tiny Comparison for survived trajectories.}
    \end{subfigure}
    \caption{Correlation matrices of 6 treatment types within clinician and proposed policies across low, medium, and high SOFA severity levels, categorized by (a) survived and (b) deceased outcomes.}
    \label{fig:policy_comparison}
\end{figure}

We further assess how our learned policy aligns with clinician behavior in multi-organ intervention contexts. Specifically, we compare the pairwise treatment correlations across six intervention types between the clinician and the proposed policy. These comparisons are stratified by SOFA-based patient severity (i.e., low, medium and high) and survival outcome (i.e., survived vs. deceased) (see Figure \ref{fig:policy_comparison}). Note that exact alignment with clinician practice is neither expected nor optimal since it is well documented that clinicians often make suboptimal decisions under uncertainty in complex disease management. However, alignment with positive outcomes (i.e. survival) and divergence in high-risk cases (i.e., non-survivors) would showcase a robust and adaptive policy. It is evident that our policy closely replicated the positive treatment synergies observed in survivors across different severity levels, while showcasing reduced correlation with deceased cohorts. This indicates that the proposed model was able to recognize effective clinical patterns and also learned to avoid suboptimal treatment combinations associated with poorer outcomes. These findings showcase the policy's ability for adaptive decision-making based on patient state severity that could lead to superior outcomes.
\subsubsection{\textbf{RQ4: How effective are the proposed dual-layer state representations and cross-agent communication mechanisms?}}
We measured the contributions of two core components of our solution: dual-layer state representations and cross-agent communications. We trained three variations of our hierarchical model: without inter-agent communications (Prop-NoC), without the proposed state representations, using raw feature vectors (Prop-NoSR), and without (Prop-NoC-NoSR). All models showed degradation in mortality rates and CWPDIS values (see Tables \ref{table:quant_eval} and \ref{table:ablation}). Compared to our full model, we see an increase of 27.2\%, 19.7\%, 33.4\% in mean mortality for Prop-NoC, Prop-NoSR, and Prop-NoC-NoSR. These results demonstrate the critical role of these components.
%in handling the task complexity. 

% \begin{table}[h]
% \small
% \caption{Off-policy evaluation metrics of the proposed model, without inter-agent communication and state representations.}
% \label{tab:ablation}
% \centering
% \begin{tabular}{|c|c|c|c|}
% \hline
% \textbf{Model}   & \textbf{ESS} & \textbf{V} & \textbf{Mortality (\%)} \\ \hline
% Prop-NoC                        & \multicolumn{1}{c|}{4775}            & 26.27          & 10.54 ± 0.27                                    \\ \hline
% Prop-NoSR                       & \multicolumn{1}{c|}{1020}            & 26.13          & 11.21 ± 0.33                                    \\ \hline
% Prop-NoC-NoSR                   & \multicolumn{1}{c|}{3633}            & 24.80          & 11.75 ± 0.50                                    \\ \hline
% \end{tabular}
% \end{table}

\begin{table}%[ht]
\caption{Off-policy evaluation metrics of the proposed model, without inter-agent communication and state representations.}
\footnotesize
\centering
\begin{tabular}{lcc}
\toprule
\textbf{Model} & \textbf{V} & \textbf{Mortality (\%)} \\
\midrule
Prop-NoC         & 18.54 & 12.65 $\pm$ 0.27 \\
Prop-NoSR        & 19.01 & 11.92 $\pm$ 0.36 \\
Prop-NoC-NoSR    & 16.18 & 13.27 $\pm$ 0.31 \\
\bottomrule
\end{tabular}

\label{table:ablation}
\end{table}

% {\color{pink} Make sure table fonts are not too small.}
\section{Conclusions}
We introduce a hierarchical multi-agent reinforcement learning framework for the complex and first-of-its-kind multi-organ treatment recommendations, setting a new benchmark in clinical decision support systems. Mimicking real world collaborative clinical settings, our solution effectively decomposes the complex treatment process into a clinically meaningful hierarchy of subtasks. Each subtask is managed by specialized agents operating within dedicated subspaces. It supports independent and cooperative agent functionality through robust inter-and intra- organ communications. Moreover, a dual-layer state representation technique is proposed to support advanced contextualization needed at multiple levels in the hierarchy. We evaluate our solution on the non-standardized and multi-dimensional sepsis treatment recommendation problem. Comprehensive quantitative and qualitative evaluation on internal and external validation datasets showed that our solution consistently outperformed baselines, significantly improving the patient survival and effectively managing task complexity. Furthermore, learned policy closely followed successful clinical treatment patterns, deviating only when beneficial, thus enhancing the reliability of the policy. The inherent flexibility and scalability of our solution allow it to be expanded to a broader range of treatments and organ, and also for complex scenarios beyond healthcare. 

% In this study, we developed a flexible, hierarchical, multi-agent reinforcement learning solution to address the complex multi-organ treatment recommendation task. Our solution effectively simplifies the high-dimensional multi-organ treatment space through the delegation of specific subtasks across multiple specialized agents operating in a decision-making hierarchy both individually and cooperatively. The cooperation among the agents were supported through both inter- and intra-organ agent communication mechanisms, allowing the model to explicitly learn relevant treatment interactions. Additionally, our solution introduces the use of unified and targeted patient state representations which contextualizes raw features to the unique physiology of each organ system, allowing for more effective learning. We evaluate our solution on non-standardized and multi-dimensional sepsis treatment recommendation. Comprehensive quantitative and qualitative evaluation demonstrated that our solution consistently outperformed baselines, improving the likelihood of patient survival and effectively managing high task complexity. Our solution is designed to be highly flexible and scalable; allowing it to be effectively extended to include a wider array of treatments and organ systems in future works. Furthermore, we believe that the proposed solution is generic and can be utilized to handle complex recommendation tasks beyond the healthcare sector. 

% \newpage
% \null
% \newpage
\balance
\bibliographystyle{ACM-Reference-Format}
\bibliography{sigkdd2026}


\newpage
\appendix
\section{Methodology}
\subsection{RL Actions:}
For each 4-hour window, we computed the total dosage for each treatment by multiplying its infusion rate with the overlapping duration, and adding any IV push volumes. Treatments with multiple drugs were converted into their standard equivalents (i.e., vasopressors into norepinepherine \cite{kotani2023updated}, S2 to fentanyl \cite{mcpherson2010demystifying}, and diuretics to furosemide \cite{konerman2022michigan}) (see Table \ref{table:treatment_summary}). Thresholds used for action discretization can be found in Table \ref{table:action_thresholds}.

\subsection{RL States:}
State variables were chosen based on clinical relevance and concurrent availability in the MIMIC-IV and AmsterdamUMC datasets (see Table \ref{table:model_features_sepsis}). The data was aggregated into 4-hourly windows using mean or sum as appropriate. Missing values were imputed using the 'last-observation carried forward' method (LOCF). Binary features were normalized to -0.5 and 0.5, and normally and log-normally distributed features to 0-1 range.

\begin{figure*}[!t]
    \centering
    \includegraphics[width=0.7\linewidth]{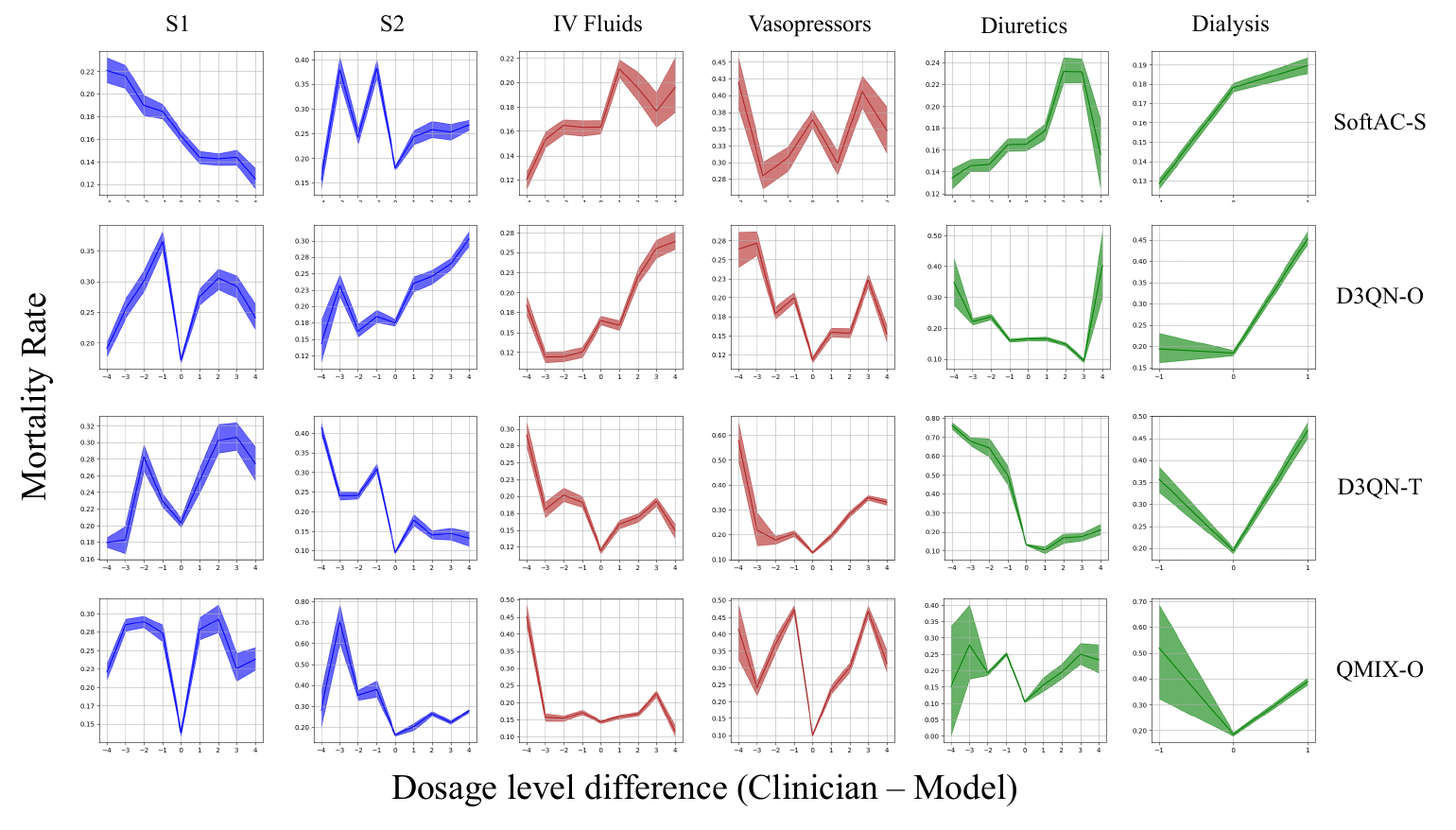}
    \caption{Baseline performances omitted due to space limitations from main text. Dosage differences (x-axis) versus mortality (y-axis) for individual agents in the hierarchy, with colors blue, red, and green denoting different organ systems.}
    \label{fig:baseline_v_plots}
\end{figure*}

% \begin{table}[]
% \centering
% \small
% \caption{Summary of treatments, clinical components, purposes, and number of samples per treatment}
% \resizebox{\columnwidth}{!}{%
% \begin{tabular}{@{}p{2cm}p{2.3cm}p{2.7cm}p{3.2cm}p{1.2cm}@{}}
% \toprule
% \textbf{Organ System} & \textbf{Treatment} & \textbf{\begin{tabular}[c]{@{}c@{}}Clinical \\ Components\end{tabular}} & \textbf{Purpose} & \textbf{Samples} \\ \midrule
% \multirow{Neurological} & S1 & Propofol & Anesthesia & 71,389 \\ \cmidrule(l){2-5} 
%  & S2 & Fentanyl, Morphine & Analgesia (pain relief) & 89,400 \\ \cmidrule(l){2-5} 
% \multirow{Cardiovascular} & IV Fluids & Crystalloids, colloids, blood products (tonicity adjusted) & Hemodynamic Support, hydration, electrolyte balance & 269,672 \\ \cmidrule(l){2-5} 
%  & Vasopressors & Norepinephrine, Epinephrine, Dopamine, Vasopressin, Phenylephrine & Increase blood pressure for improved organ perfusion; stabilize hemodynamics & 55,591 \\ \midrule
% \multirow{Renal} & Diuretics & Furosemide, Bumetanide & Reduce blood pressure and edema; maintain fluid balance & 17,279 \\ \cmidrule(l){2-5} 
%  & Dialysis & Active renal replacement therapy & Waste product removal, maintain electrolyte and fluid balance; support kidney function & 7,135 \\ \bottomrule
% \end{tabular}%
% }
% \label{table:treatment_summary}
% \end{table}

\begin{table}[]
\centering
\small
\caption{Summary of treatments, clinical components, purposes, and number of samples per treatment}
\resizebox{\columnwidth}{!}{%
\begin{tabular}{@{}p{2cm}p{2.3cm}p{2.7cm}p{3.2cm}p{1.2cm}@{}}
\toprule
\textbf{Organ System} & \textbf{Treatment} & \textbf{Clinical Components} & \textbf{Purpose} & \textbf{Samples} \\
\midrule
Neurological     & S1            & Propofol                             & Anesthesia                                              & 71,389  \\
                 & S2            & Fentanyl, Morphine                   & Analgesia (pain relief)                                 & 89,400  \\
\midrule
Cardiovascular   & IV Fluids     & Crystalloids, colloids, blood products (tonicity adjusted) & Hemodynamic support, hydration, electrolyte balance     & 269,672 \\
                 & Vasopressors  & Norepinephrine, Epinephrine, Dopamine, Vasopressin, Phenylephrine & Increase blood pressure; stabilize hemodynamics   & 55,591  \\
\midrule
Renal            & Diuretics     & Furosemide, Bumetanide               & Reduce BP and edema; maintain fluid balance              & 17,279  \\
                 & Dialysis      & Active renal replacement therapy     & Waste removal, support kidney function                  & 7,135   \\
\bottomrule
\end{tabular}%
}
\label{table:treatment_summary}
\end{table}

\begin{table}[ht]
\centering
\scriptsize
\caption{Action level thresholds (4-hourly doses)}
\setlength{\tabcolsep}{2.5pt} % Reduce column spacing
\begin{tabular}{p{3cm}ccccc}
\toprule
\textbf{Treatment} & \textbf{0} & \textbf{1} & \textbf{2} & \textbf{3} & \textbf{4} \\
\midrule
IV Fluids (ml) & 0 & 0.00--56.17 & 56.17--227.50 & 227.50--530.93 & $>$530.93 \\
Vasopressor (\textmu g/kg) & 0 & 0.00--9.40 & 9.40--20.66 & 20.66--44.42 & $>$44.42 \\
S1 (mg/kg of propofol) & 0 & 0.00--3.40 & 3.40--6.01 & 6.01--9.53 & $>$9.53 \\
S2 (\textmu g/kg of fentanyl) & 0 & 0.00--0.65 & 0.65--2.08 & 2.08--4.30 & $>$4.30 \\
Diuretics (mg of furosemide) & 0 & 0.00--20.00 & 20.00--160.00 & 160.00--902.10 & $>$902.10 \\
Dialysis & Off & On & --- & --- & --- \\
\bottomrule
\end{tabular}
\label{table:action_thresholds}
\end{table}

% \subsubsection{RL Reward:}
% In addition to the mortality based terminal reward ($+R/-R$), in line with the literature, we use a clinically guided intermittent reward function $R_{im}(s_{t},s_{t+1})$, calculated from the transition from $s_{t}$ to $s_{t+1}$ based on immediate impact on person's health post-treatment \cite{raghu2017deep}.

% \vspace{1.5em}
% \noindent\textbf{\normalsize Training Process:} \\[-0.3em]
\subsection{Training Process}
We describe the overall training process in Algorithm \ref{alg:hmarl_training}.
\begin{algorithm}[H]

{\scriptsize
\caption{Generalized Training Process for HMARL Treatment Dynamic Treatment Recommender Solution}
\label{alg:hmarl_training}

\begin{algorithmic}[1]

\REQUIRE 
\item[] $W$: a set of organ systems \COMMENT{i.e., \{Neu, Car, Ren\}} 
\item[] $S$: learned state representations, $S = \{s_t^{Rt}, s_t^o \mid \forall o \in W\}$ 
\item[] $a_c$: clinician's action discretized into $D$ dosage levels
\item[] $r$: reward signal
\item[] $A_{Rt}, A_o$: action spaces for root and organ-specific agents

\STATE \textbf{Initialize:} $Q(s, a)$ for:
\STATE \hspace{2em} Root agent: $M_{Rt}$
\STATE \hspace{2em} Single-organ agents: $M_o$ for all $o \in W$
\STATE \hspace{2em} Treatment-level agents: $M_o^{treat}$ for all $o \in W, \forall treat \in A_o$
\STATE \hspace{2em} Mixed-treatment agents: $M_o^{Mix}$ for all $o \in W$
\STATE \hspace{2em} Mixed-organ agents: $M_{OMix}$

\STATE \textbf{Step 1: Train the Root Agent $M_{Rt}$}
\STATE Input: $s_t^{Rt}$ \COMMENT{Unified state representations}
\STATE Output: $a \in A_{Rt}$ where $A_{Rt} = \{A_{Rt}^0, A_{Rt}^{OMix}\} \cup \bigcup_{o \in W} A_{Rt}^o$

\STATE \textbf{Step 2: Train Single-Organ System Agents}
\FOR{$o \in W$}
    \STATE Train $M_o(s_t^o, a)$ on $a_c \in A_o$
    \STATE Output: $a_o^{master} \in \bigcup_{i=1}^{|A_o|} a_o^{master_i}$
    
    \FOR{$treat \in A_o \setminus A_o^{Mix}$}
        \STATE Train $M_o^{treat}(s_t^o, a)$ on $a_c \in A_o \setminus A_o^{Mix}$
        \STATE Output: $a \in \bigcup_{d \in D} a_o^{treat,d}$
    \ENDFOR
    
    \STATE Train $M_o^{Mix}(s, a)$ on $a_c \in A_o^{Mix}$, where:
    \STATE $s = \text{concat}(s_t^o, a_o^{treat} \mid \forall treat \in A_o \setminus A_o^{Mix})$
    \STATE Output: $a \in \bigcup_{(treat, d) \in |A_o| \times D} a_o^{treat,d}$
\ENDFOR

\STATE \textbf{Step 3: Retrieve Single-Organ System Actions}
\FOR{$o \in W$}
    \STATE Select $a_o^{master}$ from $M_o$
    \STATE Retrieve $G_o = \{a_o^{treat_1}, \dots, a_o^{treat_n}\}$, where $n = |A_o|$
\ENDFOR

\STATE \textbf{Step 4: Train the Mixed-Organ Agents using QMIX}
\STATE Agents: $M_{OMix}(s, a)$ and $M_{OMix}^o$ for all $o \in W$
\STATE Input: $s = \big[ s_t^y, G_y \mid \forall y \in W \big]$
\STATE Output: Mixture of treatments across organ systems: $A_t^{OMix}$

\STATE \textbf{Step 5: Phase-2 Training}
\STATE See Equation (1) for detailed updates

\end{algorithmic}
}
\end{algorithm}

% In addition to the mortality based terminal reward ($+R/-R$), in line with the literature, we use a clinically guided intermittent reward function $R_{im}(s_{t},s_{t+1})$, calculated from the transition from $s_{t}$ to $s_{t+1}$ ({\color{pink}(CITE)}). The function penalizes increases in SOFA or lactate levels, and non-zero SOFA scores that remain unchanged. The contribution of each of these components is controlled by tunable parameters $C_0$, $C_1$, $C_2$, these parameters also serve to ensure that the magnitude of the intermittent rewards do not overpower the terminal reward.
% {\color{pink} Maybe we can omit these equations and use the citation.}

% \begin{align}
% R_{\text{im}}(s_t, s_{t+1}) &= C_0(s_{t+1}^{\text{SOFA}} = s_t^{\text{SOFA}} \land s_{t+1}^{\text{SOFA}} > 0) \notag \\
% &\quad + C_1(s_{t+1}^{\text{SOFA}} - s_t^{\text{SOFA}})\notag \\ 
% &\quad + C_2 \tanh(s_{t+1}^{\text{lactate}} - s_t^{\text{lactate}})
% \label{eq:rim}
% \end{align}

% \begin{equation}
% R(s_t, s_{t+1}) =
% \begin{cases}
% +R_r, & \text{if } s_{t+1} \in S_T \cap S_{\text{sur}} \\
% -R_r, & \text{if } s_{t+1} \in S_T \cap S_{\text{dec}} \\
% R_{\text{im}}(s_t, s_{t+1}), & \text{if } s_{t+1} \notin S_T
% \label{eq:r}
% \end{cases}
% \end{equation}
\begin{table}[]
\centering
\small
\caption{List of model features}
\begin{tabular}{p{2.5cm}p{5.5cm}}
\toprule
\textbf{Category} & \textbf{Feature Name} \\ \midrule
Demographics (8) &
  Age, Shock index, SOFA, GCS, Weight, SIRS, Gender, Readmission \\ \midrule
Vital signs (10) &
  HR, SBP, MBP, DBP, Resp, Temp., PaCO2, PaO2, PaO2/FiO2 ratio, SpO2 \\ \midrule
Lab values (21) &
  Albumin, pH, Calcium, Glucose, Hb, Magnesium, WBC, Creatinine, Bicarbonate, Sodium, Lactate, Chloride, Platelets, Potassium, PTT, AST, ALT, BUN, Ionised calcium, Total bilirubin, Base excess \\ \midrule
Output events (2) &
  Fluid output (4 hourly), Total output \\ \midrule
Ventilation \& others (3) &
  Mechanical ventilation, FiO2, Timestep \\ \bottomrule
\end{tabular}
\begin{tablenotes}
  \setlength\labelsep{0pt}
      \item 
      \tiny{Abbreviations- PTT: Partial Thromboplastin Time; SIRS: Systemic Inflammatory Response Syndrome; ICU: Intensive care unit; WBC: White blood cell; Temp.: Temperature; GCS: Glasgow Coma Scale; Resp.: Respiratory rate; HR: Heart rate; SBP: Systolic blood pressure; MBP: Mean blood pressure; DBP: Diastolic blood pressure; Hb: Hemoglobin}
\end{tablenotes}
\label{table:model_features_sepsis}
\end{table}
\section{Experiments}

\subsection{Mortality vs. Difference in Recommended Dosage:} See Figure \ref{fig:baseline_v_plots} for remaining plots omitted from main text.

\subsection{Ethical Considerations:} Similar to RL based Clinical Decision Support Systems (CDSSs) proposed in the literature for various diseases including sepsis management \cite{komorowski2018artificial,raghu2017deep,saria2018individualized}, our solution is a human in the loop CDSS; where clinicians could utilize the data driven decisions provided by the intelligent agents as auxiliary inputs before making the final recommendations. In addition to the final action recommendations from the CDSS, the $Q(s,a)$ values available for all possible actions ($a$) could be used to quantify and compare the quality of the available treatment options for a given patient, with respect to his long term survival. The proposed solution allows clinicians to collaborate with the intelligent agent that combines the experiences of successful clinical decisions in the past, but leaves the full control of the final decision to the clinician. Furthermore, in line with the literature, all models were trained and evaluated using publicly available retrospective data, and no clinical trials were conducted. Therefore, the proposed solution does not raise ethical concerns.

\subsection{Reproducibility and Model Parameters:}
Source codes along with tuned parameters and architectures are uploaded with the submission and available online as an anonymous repository\footnote{https://anonymous.4open.science/r/HMARL-PyTorch-022B/}. All models are tested on the same holdout set. We experimentally set $\gamma$ to 0.99 and $k$ to 8. Data for the sepsis cohort can be replicated using the provided SQL scripts and Python codes.

\end{document}